\documentclass[10pt, conference]{IEEEtran}
\usepackage{caption,subfig}
\usepackage{textcomp}
\usepackage{epsfig,graphicx, graphics}
\usepackage{xcolor}
\usepackage{amsfonts,amsmath,amssymb}
\usepackage{fixltx2e} 
\usepackage{booktabs}
\makeatletter
\let\@currsize\normalsize
\makeatother
\usepackage{setspace}
\usepackage[ruled,vlined]{algorithm2e}
\usepackage{soul} 
\usepackage{tablefootnote}
\usepackage{multirow}
\usepackage{hyperref}
\usepackage{url}
\usepackage{wrapfig}

\begin{document}

\title{Using Deep Convolutional Networks for Occlusion Edge Detection in RGB-D Frames}

\author{\IEEEauthorblockN{Soumik Sarkar\IEEEauthorrefmark{1},
Vivek Venugopalan\IEEEauthorrefmark{2}, Kishore Reddy\IEEEauthorrefmark{2}, Michael Giering\IEEEauthorrefmark{2}, Julian Ryde\IEEEauthorrefmark{2} and Navdeep Jaitly\IEEEauthorrefmark{3}}
\IEEEauthorblockA{\IEEEauthorrefmark{1}Iowa State University, Ames, IA, USA Email:\{soumiks@iastate.edu\}}
\IEEEauthorblockA{\IEEEauthorrefmark{2}United Technologies Research Center, East Hartford, CT 06108, USA, \\Email: {\{gierinmj, reddykk, venugov, rydejc\}@utrc.utc.com}}
\IEEEauthorblockA{\IEEEauthorrefmark{3}Google Inc., Mountain View, CA, Email:\{ndjaitly@google.com\}}}

\maketitle
\begin{abstract}
Occlusion edges in images which correspond to range discontinuity in the scene from the point of view of the observer are an important prerequisite for many vision and mobile robot tasks. Although occlusion edges can be extracted from range data, extracting them from images and videos is challenging and would be extremely beneficial for a variety of robotics based applications. We trained a deep convolutional neural network (CNN) to identify occlusion edges in images and videos with both RGB-D and RGB inputs. The use of CNN avoids hand-crafting of features for automatically isolating occlusion edges and distinguishing them from appearance edges. Other than quantitative occlusion edge detection results, qualitative results are provided to demonstrate the trade-off between high resolution analysis and frame-level computation time which is critical for real-time robotics applications.
\end{abstract}

\section{Introduction}\label{sec:intro}
Occlusion edge detection is a fundamental capability of computer vision systems as is evident from the number of applications and significant attention it has received
~\cite{jacobson2012online,ayvaci2011detachable,sargin2009probabilistic,marshall1996occlusion,stein2009occlusion}.
Occlusion edges are useful for a wide array of tasks including object recognition, feature selection, grasping, obstacle avoidance, navigating, path-planning,  localization, mapping, stereo-vision and optic flow. In addition to numerous applications, the concept of occlusions edges is supported by the human visual perception research~\cite{wagemans2012century} where it is referred to as figure/ground determination.
Once occlusion boundaries have been established, depth order of regions become possible \cite{sundberg2011occlusion,smith2004layered} which aids navigation, simultaneous localization and mapping (SLAM) and path planning.
Occlusion edges help image feature selection by rejecting features generated from regions that span an occlusion edge. As these are dependent on viewpoint position, removing these variant feature saves on further processing and increases recognition accuracy~\cite{gil2010comparative}.
In many object recognition problems, the shape of the object is better for recognition rather than its appearance, which can be easily dramatically altered e.g., by painted objects, camouflage and people wearing different clothes. However, shape determination is not the approach for state-of-the-art SIFT based object recognition algorithms.
Furthermore, knowledge of occlusion edges helps with stereo vision~\cite{belhumeur1992bayesian} and optic flow algorithms~\cite{sundberg2011occlusion}.
In robotics, geometric edges of objects demarcate their spatial extents helping with grasping, manipulation as well as maneuvering through the world without collision and therefore, knowledge of occlusion edges is essential.

In this context, this paper evaluates the efficacy of Deep Learning tools~\cite{BO11} for the task of occlusion edge detection. Recently, this class of techniques have emerged as the top performing machine learning tool for various tasks such as object recognition~\cite{KSH12}, speech recognition~\cite{HDY12}, denoising~\cite{VLB08}, hashing~\cite{SH09} and data fusion~\cite{SS14}. While Deep Neural Networks (DNN) pre-trained using Deep Belief Networks (DBN)~\cite{RB08,HS06} perform quite well in most data types, deep Convolutional Neural Networks~\cite{KSB10} have been shown to be most suited for images. The better performance is primarily attributed to the preservation of local structures (i.e., localized pixel dependencies) by CNN as opposed to DBN-DNN (where, typically layers are fully connected bipartite graphs). The occlusion edge detection task can logically be conceived as a two step process: identifying edges in an image followed by  distinguishing between occlusion and appearance edges. Therefore, deep neural networks are particularly interesting for this problems as they extract hierarchical features (features of features) from data and visualization of intermediate optimized filters~\cite{KSH12} show that edge type features are very common. It also should be noted that such an approach eliminates the need for complicated hand-crafting of features that is commonly done in many current approaches. Due to availability of GPUs and recent advancements in the algorithmic/implementation side, large CNNs can be learnt without significant overfitting from high volume of data for complex problems~\cite{KSH12}. In fact, the CNN model size (depth and breadth) can be optimized iteratively for a certain problem. Often however, memory of the implementing GPU becomes the bottle-neck.

In this paper, the main contributions are: (i) formulation of an occlusion edge detection problem as a classification of center-pixels of an image patch with RGB-D channels (ii) performance evaluation of CNN with various input information, namely RGB-D single time frame and RGB single time frame (iii) fusion of patch predictions to generate frame-wide occlusion edges. The study uses a publicly available benchmark RGB-D data set captured with moving camera in an indoor environment by the Computer Vision group at Technische Universität München (TUM)~\cite{SEE12}. The optimized and hardware-accelerated CNN implementation has been done on NVIDIA K-40 GPU. The paper is organized in seven sections including the introduction. The problem formulation along with the data set description is provided in Section~\ref{sec:problem}. While Section~\ref{sec:CNN} provides the details of architecture and training parameters for the CNN, testing and post-processing are discussed in Section~\ref{sec:testing}. Various experiments with corresponding quantitative results are provided in Section~\ref{sec:exp} and qualitative observations are articulated in Section~\ref{sec:qual}. Finally, the paper is summarized and concluded with future research directions in Section~\ref{sec:con}.

\section{Problem Formulation and Generating the Training data}\label{sec:problem}

\begin{figure*}[htbp]
	\centering
    \includegraphics[width=0.32\linewidth]{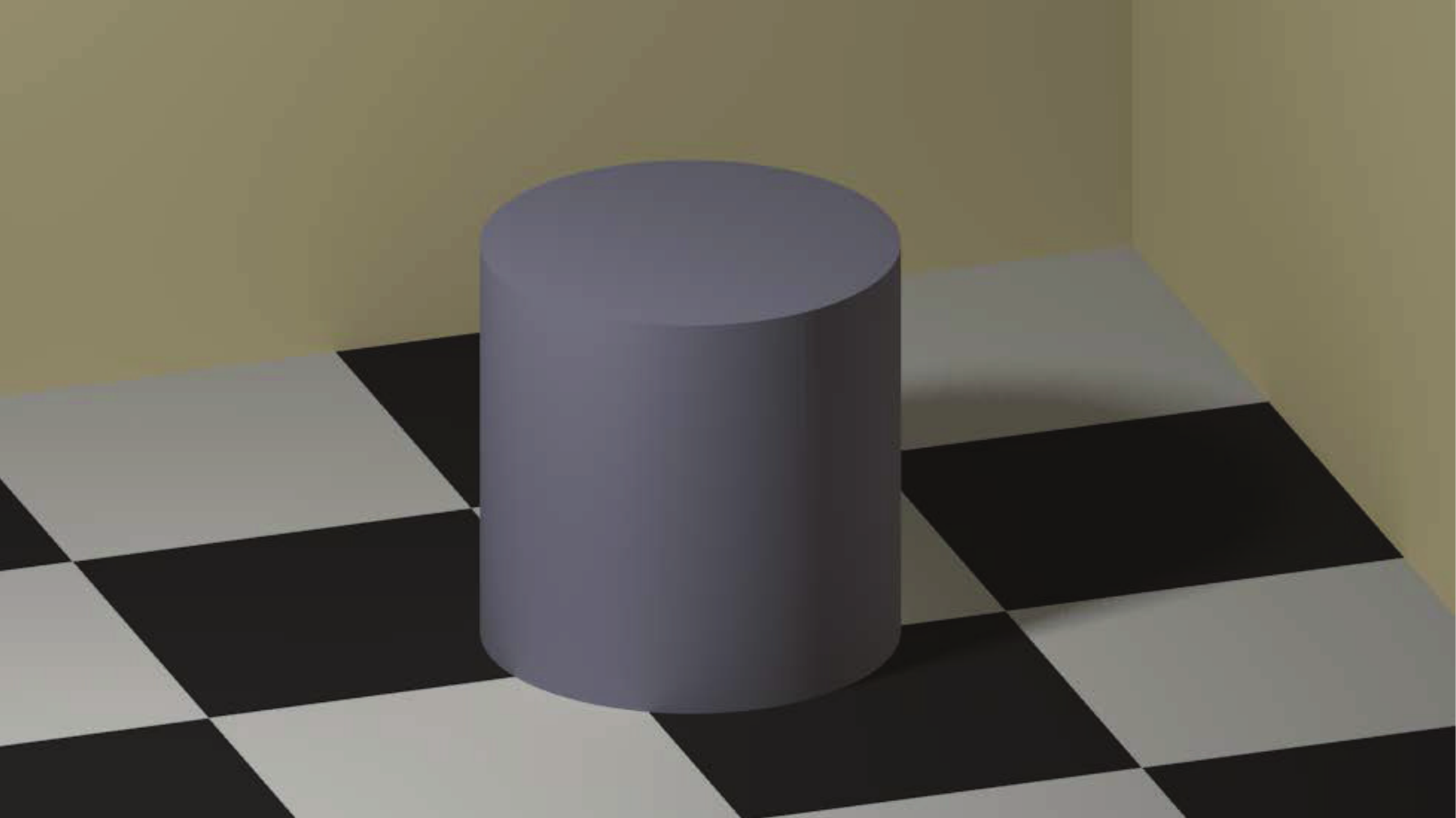}
    \includegraphics[width=0.32\linewidth]{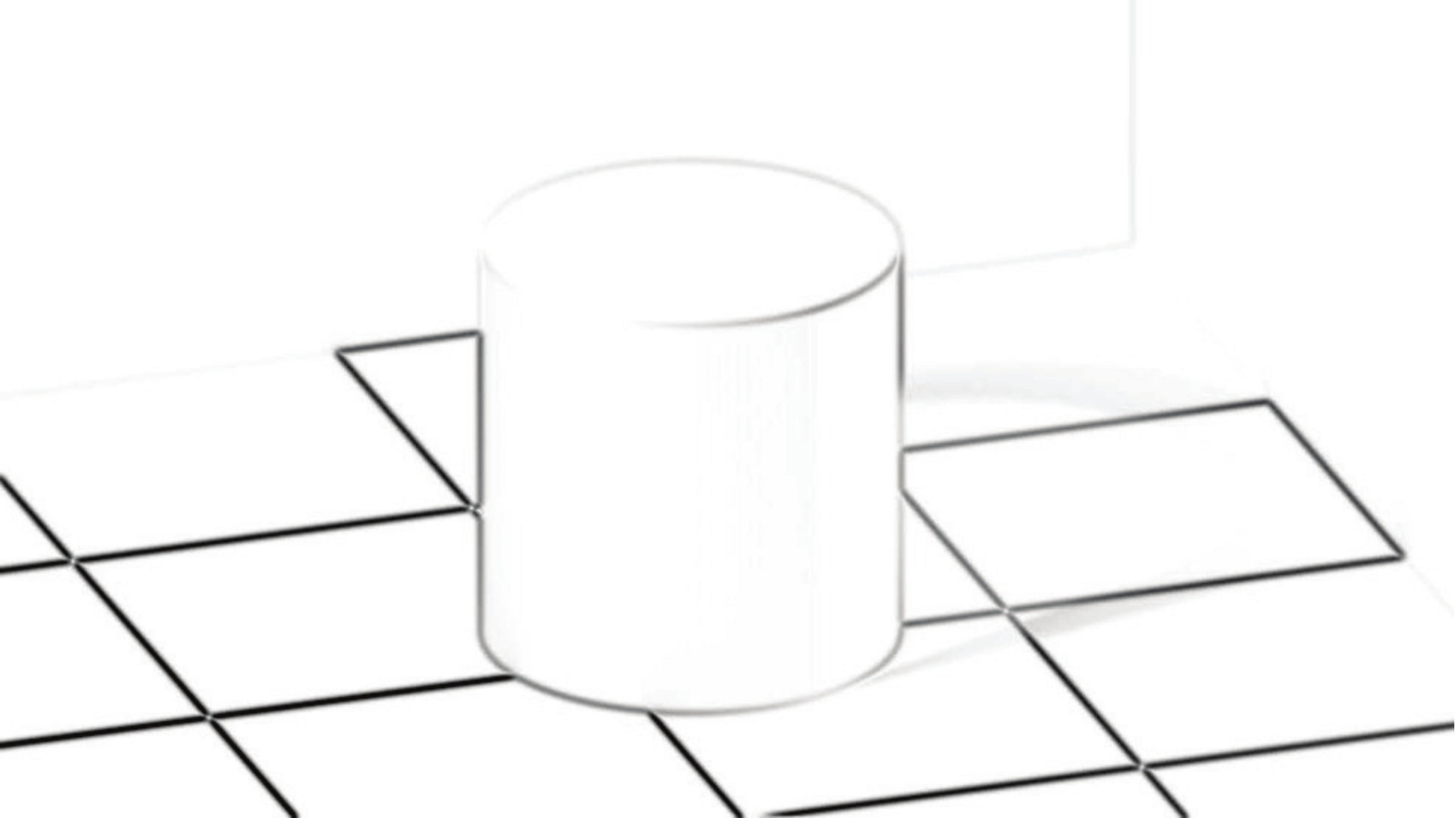}
    \includegraphics[width=0.32\linewidth]{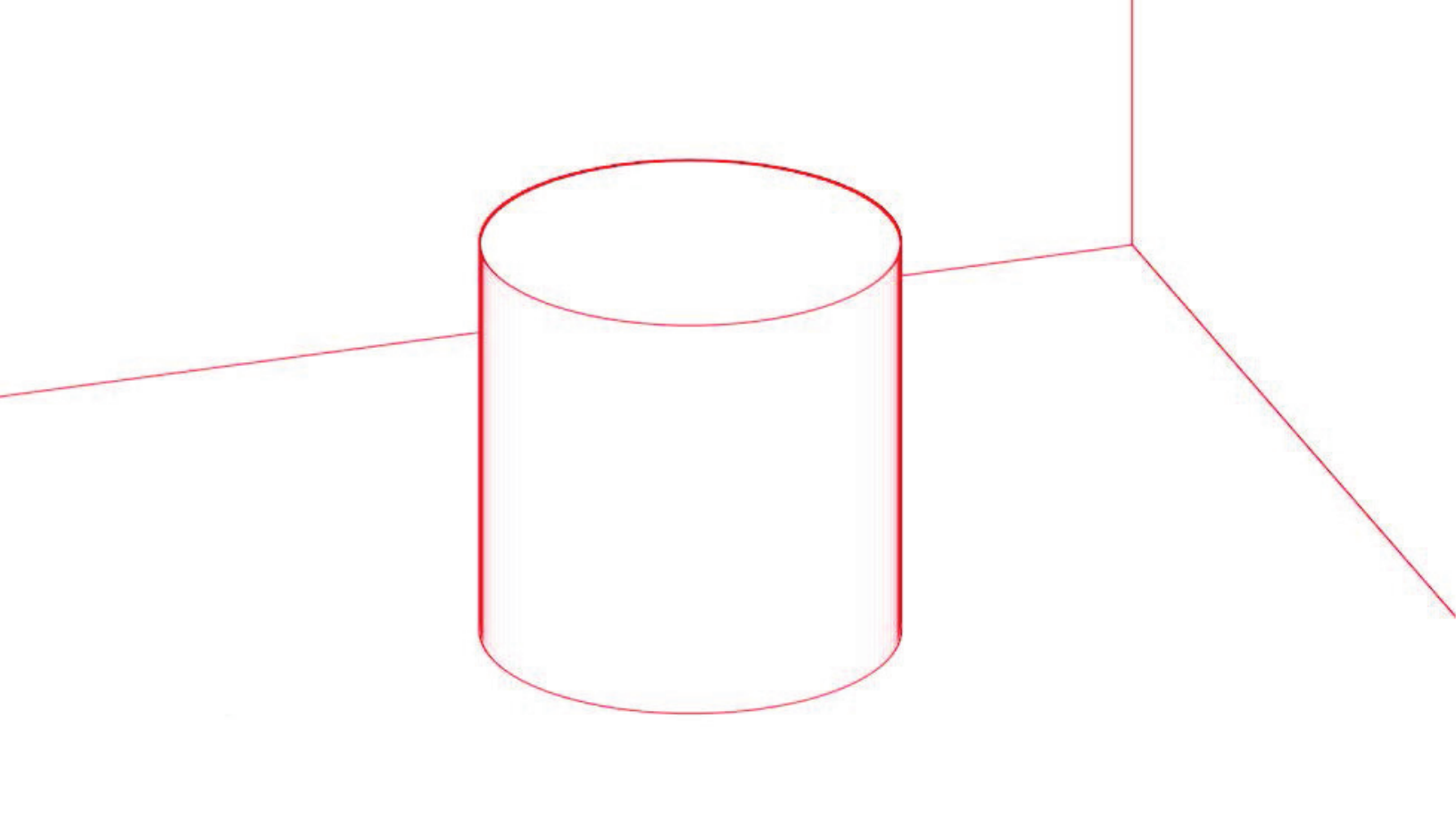}
    \caption{Image with associated edges due to appearance and due to geometry.}
    \label{fig_edge_types}
\end{figure*}

In general, it is difficult to define edge pixels rigorously. In an image, edges manifest along paths of high contrast and are due to four main reasons: (i) texture change, i.e., abrupt change in surface color, (ii) lighting change, i.e., sharp shadows, (iii) range discontinuity, i.e., abrupt change in distance from the observer and (iv) surface normal change, e.g., intersection of two planes. It is important to appreciate the distinction in the causes of image edges. Texture change and illumination edges are not observed by 3D sensors. Therefore, the remaining geometric edge types are range discontinuities and abrupt surface normal changes. Surface normal changes are pose invariant, however edges due to range discontinuities can vary with observer position. These surface normal and range discontinuities are illustrated in the last image of Fig.~\ref{fig_edge_types}. The cylinder sides in Fig.~\ref{fig_edge_types} are examples of range discontinuities. The position of these edges varies in 3D space as the position of the observer shifts whereas the cylinder rim edge position is consistent regardless of observer position. For use in mapping, the following characteristics is desired from extracted edge voxels: they should be generally invariant to rotation and translation, and they should be helpful in terms of constraining pose. Therefore, in this study the focus is on identifying the third and fourth type of edges.

Traditional approaches for detect geometric edges in 3D data include a keypoint detector based on a 3D extension of the Harris corner operator in the Point Cloud Library \cite{rusu2011_icra_PCL}. This detector operates on local normals of points. A related approach for selecting interest points on 3D meshes was introduced in \cite{sipiran2011harris}.
In principle, this study is similar to a recent work on indoor scene segmentation~\cite{CFN13}. However, this study focuses on if only occlusion edges can be isolated using CNNs and also if reasonable performance can be achieved without using the depth channel of the RGB-D data. As mentioned earlier, this paper uses a benchmark RGB-D data set the Computer Vision group at Technische Universität München (TUM). The data set contains RGB and depth images of a Microsoft Kinect sensor that was recorded at full frame rate (30 Hz) and sensor resolution $640 \times 480$ by moving camera in an indoor environment. The occlusion edge detection problem is formulated as a classification problem and the procedure of generating training data is provided in the following subsection.

\subsection{Training data}\label{sec:training}
\begin{figure*}[htbp]
    \centering
    \includegraphics[width=0.32\linewidth]{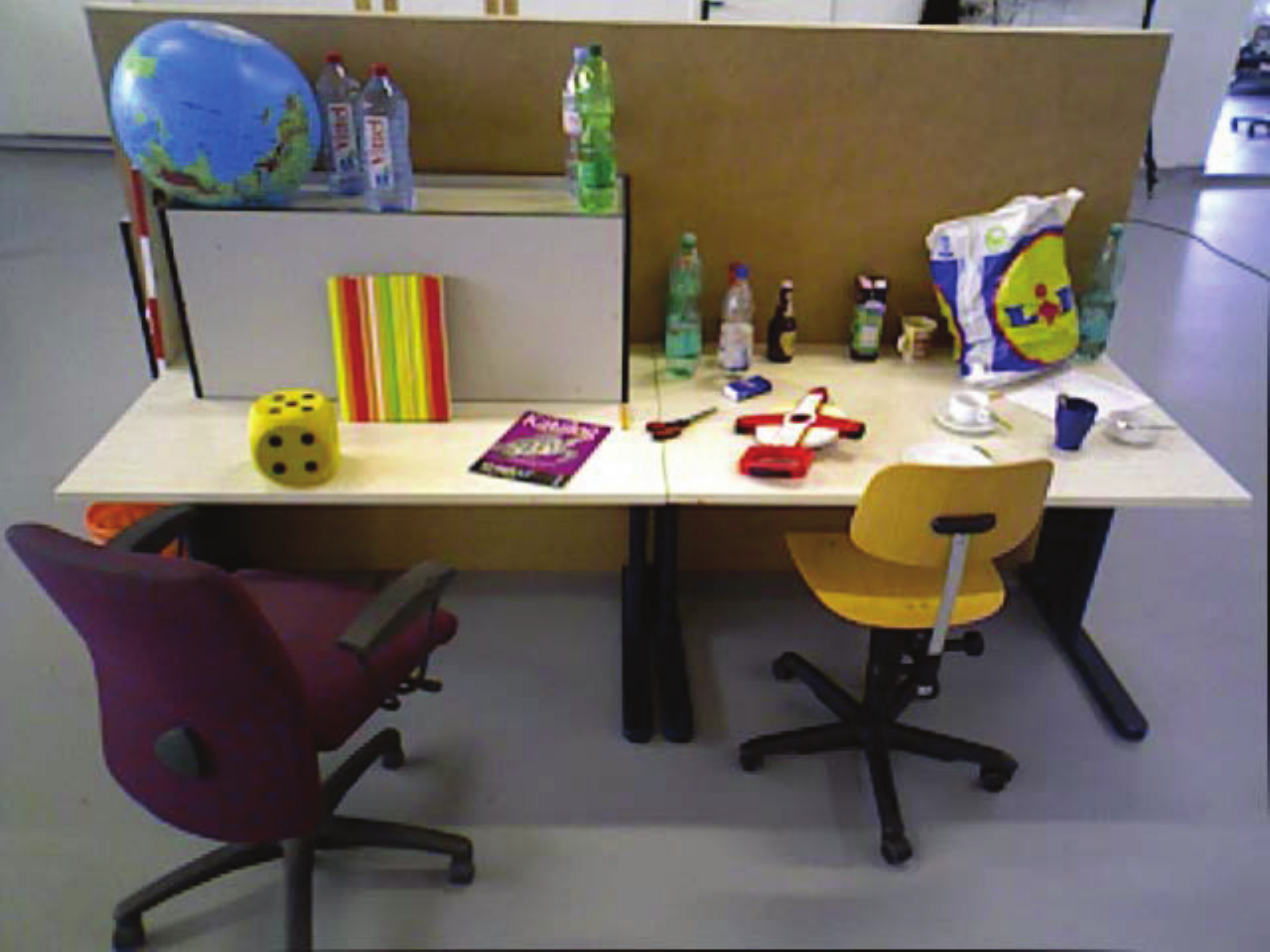}
    \includegraphics[width=0.32\linewidth]{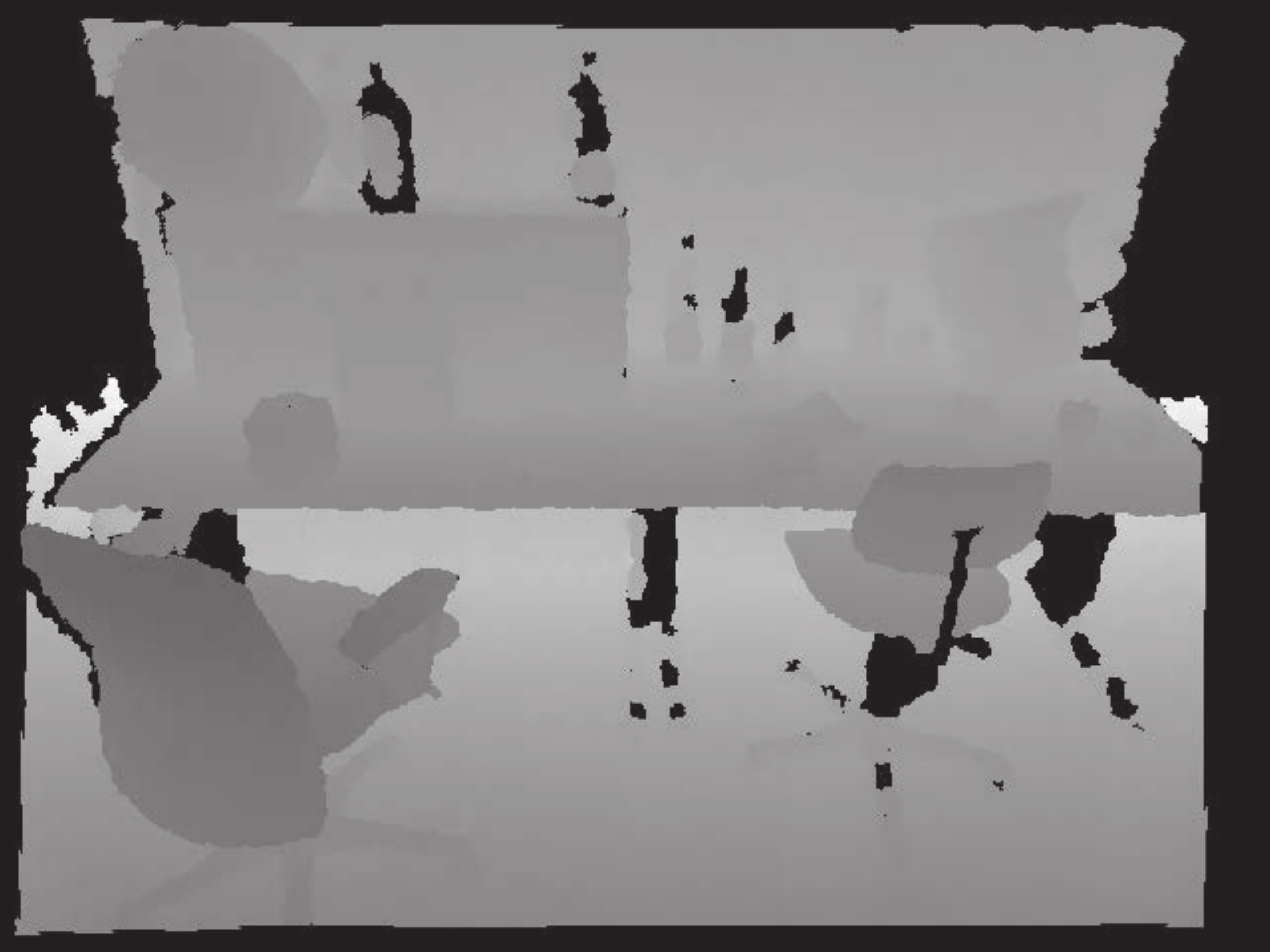}
    \includegraphics[width=0.32\linewidth]{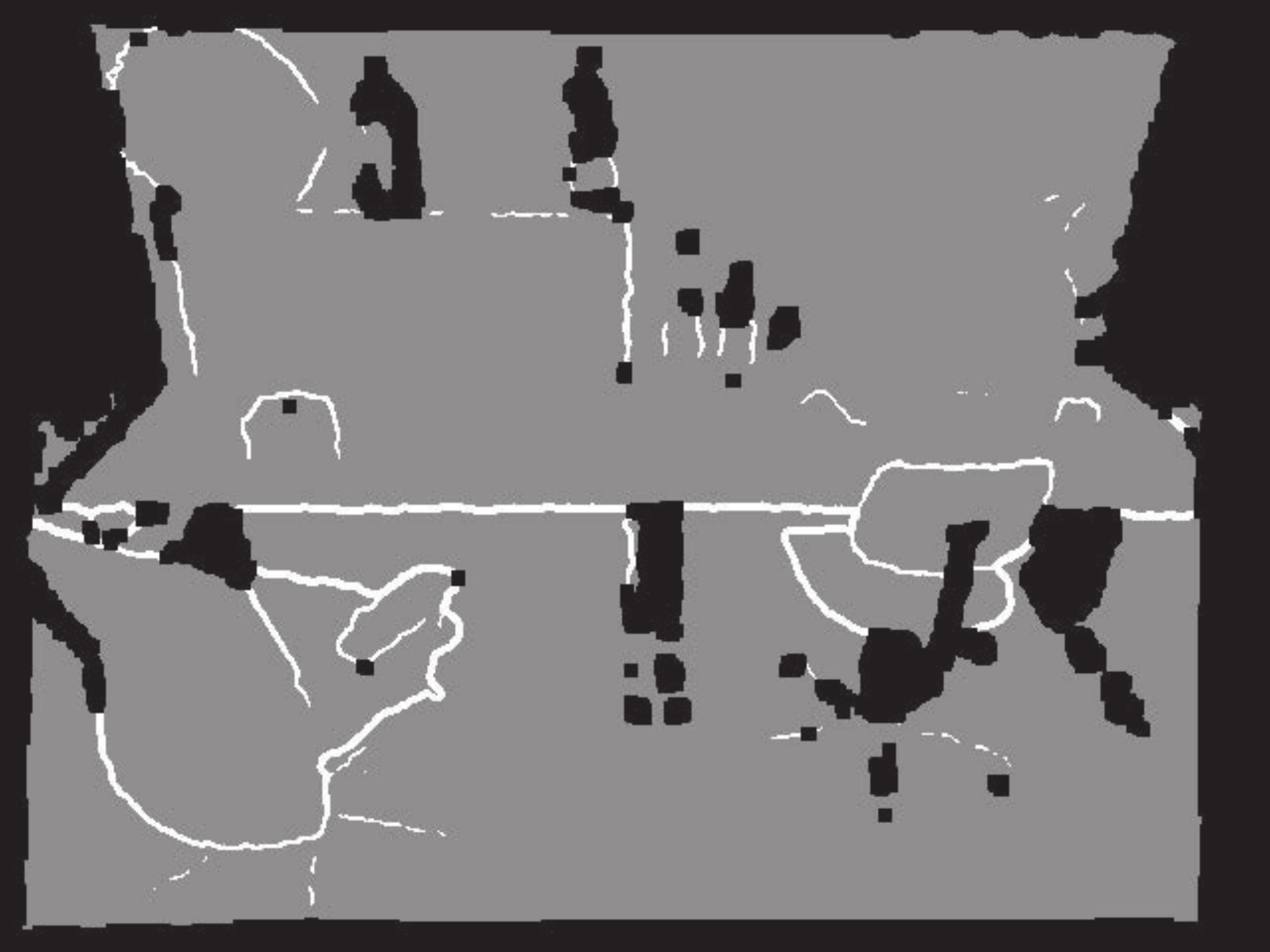}
    \caption{Example RGB, depth and classification frames from the training
        data generation procedure. In the classification frame gray signifies
        no edge, occlusion edges are white and black is for no or unreliable
        data.}
    \label{fig_rgb_edge}
\end{figure*}

The occlusion edge information is largely present in the depth (D) channel of an RGB-D frame. Therefore, occlusion edge label for a pixel, i.e., the ground truth can be automatically determined using the depth channel data. The label generation procedure is illustrated in Fig.~\ref{fig_rgb_edge}. From left to right, the three plates in the figure shows an example RGB frame, the corresponding D channel data and classification frame generated using a simple thresholding only on the depth data. Other than gray (signifying no edge) and white (signifying occlusion edges) colors, the black color can be seen in the classification frame. This signifies bad depth measurements due to presence of absorbing surface or larger than maximum distance allowed between the sensor and the surface.


\begin{figure*}[htbp]
  \centering\vspace{0pt}
  \includegraphics[width=1\textwidth]{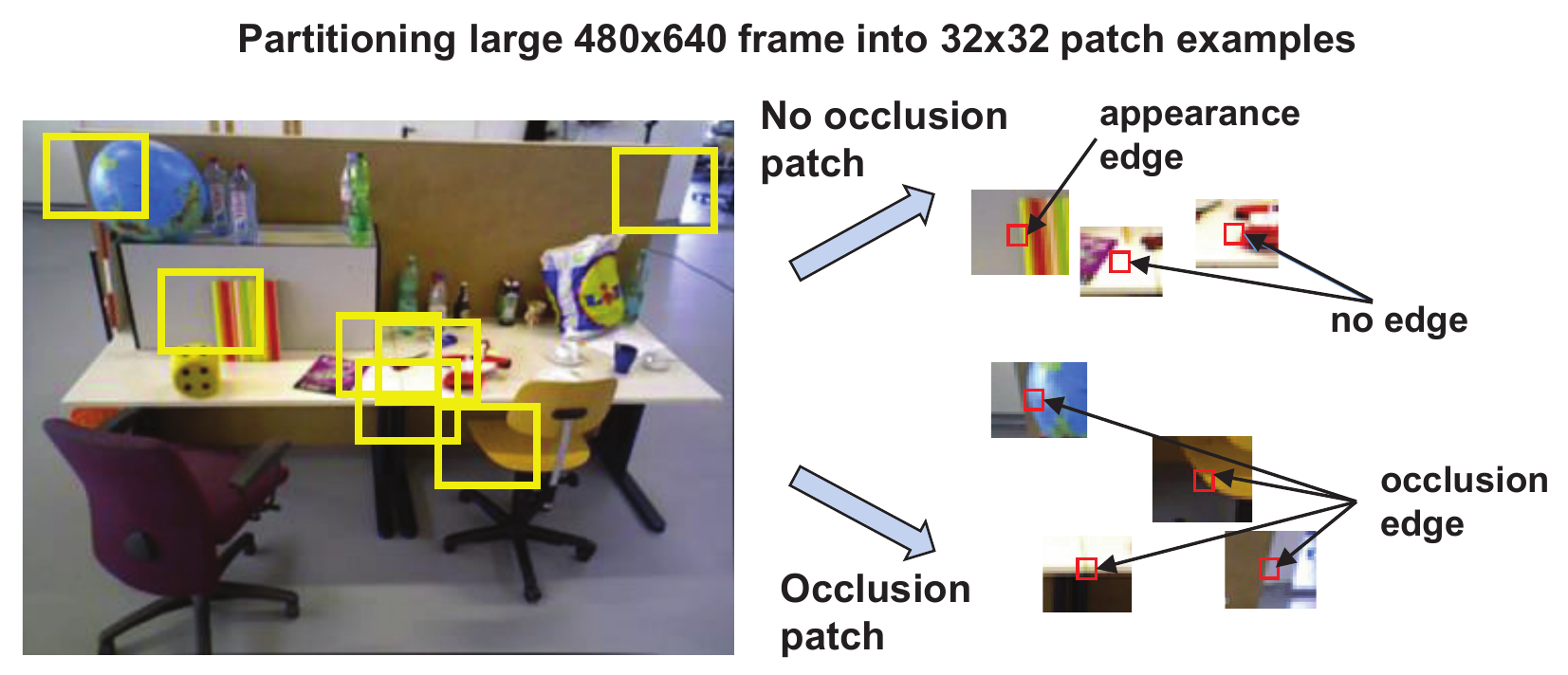}\vspace{-0pt}
  \caption{\textit{Generation of training data $32 \times 32$ patches from original $480 \times 640$ frames and labeling based on center-pixels}}\vspace{0pt}
  \label{fig:training}
\end{figure*}

The RGB-D data set was collected using a camera motion along a certain trajectory in an indoor environment. The trajectory is divided into disjoint training and testing sections so that the trained model can be tested using previously unseen data. The frames in the RGB-D data set are $480 \times 640$ in size. In order to create training examples for the Convolutional Neural Network (CNN), $32 \times 32$ patches are extracted from the large frames in the training section. The training label for each patch is determined by the pixels located at the center~\cite{YGS02,KBC12}. As illustrated in Fig.~\ref{fig:training}, if majority of the pixels ($2 \times 2$ in this case) at the center of a $32 \times 32$ patch contains occlusion edges, the patch is labeled as an \textbf{Occlusion} patch. On the other hand, if center pixels contain appearance edges or no edge, corresponding patch is labeled as a \textbf{No Occlusion} patch. Patches with considerable number of bad or unlabeled pixels are pre-filtered and not used for training.

\section{CNN Architecture and Model Learning}\label{sec:CNN}

\begin{figure*}[htbp]
  \centering\vspace{0pt}
  \includegraphics[width=1\textwidth]{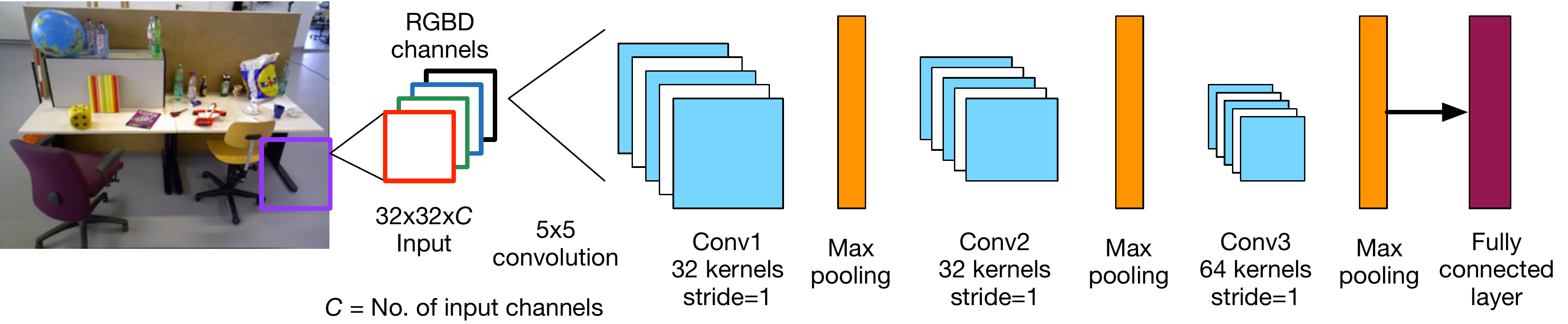}\vspace{-0pt}
  \caption{\textit{Illustration of Convolutional Neural Network (CNN) architecture used for Occlusion Edge classification}}\vspace{0pt}
  \label{fig:CNN}
\end{figure*}

The architecture of the Convolutional Neural Network (CNN) used in this paper is illustrated in Fig.~\ref{fig:CNN}. The CNN has three pairs of convolution-pooling layers followed by softmax output layer~\cite{}. This section articulates details of those layers as well as various hyper-parameters used for model learning.

\textbf{\textit{Description of layers:}} As described in Section~\ref{sec:training}, $32 \times 32$ patches were used as data for the CNN in this study. Depending on the experiment, different number channels are used for the input data. For example, while $4$ channels were used for single (time) frame RGB-D data, $3$ channels were used for a RGB frame. More detailed description of various experiments will be provided in Section~\ref{sec:exp}. The layer size parameters here correspond to the RGB-D experiment with $4$ channels. The first convolutional layer uses $32$ filters (or kernels) of size $5 \times 5 \times 4$ with a stride of $1$ pixel and padding of $2$ pixels on the edges. The CNN is configured with Rectified Linear Units (ReLUs), as they train several times faster than their equivalents with $\tanh$ connections [\cite{Nair2010Rectified-}].
A two-fold sub-sampling or pooling layer follows the convolutional layer that generates the input data (of size $16 \times 16 \times 32$) for the second convolutional layer. This layer uses $32$ filters of size $5 \times 5 \times 32$ with a stride of $1$ pixel and padding of $2$ pixels on the edges. A second pooling layer with the same specification as the first one is used after that to generate input with size $8 \times 8 \times 32$ for the third convolutional layer that uses $64$ filters of size $5 \times 5 \times 32$ with same stride and padding strategies as before. The third pooling layer also has the same configuration as the two before it and leads to a softmax output layer with two labels corresponding to \textbf{No Occlusion} and \textbf{Occlusion} classes.

\textbf{\textit{Hyper-parameters:}} The CNN described above was trained using stochastic gradient descent with a mini-batch size of $100$ examples. Although biases of convolutional layer neurons were initialized with constant values zero, weights of the neurons were initialized with zero-mean Gaussian distributions with standard deviations as: $0.0001$ for first, $0.01$ for second and $0.01$ for third convolutional layer. Interestingly, the network performed better with a comparatively larger initialization of the weight standard deviation ($0.3$) for the output layer. The learning rate and momentum used for all the convolutional layers and for all training epochs were $0.001$ and $0.9$ respectively. Finally, $L2$-regularizers were used for all convolutional layers as well with weight $0.001$. No dropout was used for model training in this study.

\textbf{\textit{Training with GPU:}} The NVIDIA Kepler series K40 GPUs are FLOPS/Watt efficient and are being used to drive real-time image processing capabilities. The Kepler series GPU consists of a maximum of 15 Streaming Execution (SMX) units and up to six 64-bit memory controllers. Each SMX unit has 192 single-precision CUDA cores and each core comprises of fully pipelined floating-point and integer arithmetic logic units. The K40 GPUs consist of 2880 cores with 12 GB of on-board device memory (RAM). Deep Learning applications have been targeted on GPUs previously in~\cite{KSH12} and these implementations are both compute and memory bound. Stacking of the channels for the RGB and the RGBD experiments result in a vector of $32 \times 32 \times 3$ and $32 \times 32 \times 4$ respectively, which is suitable for the Single Instruction Multiple Datapath (SIMD) architecture of the GPUs. At the same time, the training batch size caches in the GPU memory, so the utilization of the K40 GPU's memory is very high. This also results in our experiments to run successfully on a single GPU instead of partitioning the different layers over multiple GPUs. 

\section{Testing and Post-processing}\label{sec:testing}
\begin{figure*}[htbp]
  \centering\vspace{0pt}
  \includegraphics[width=1\textwidth]{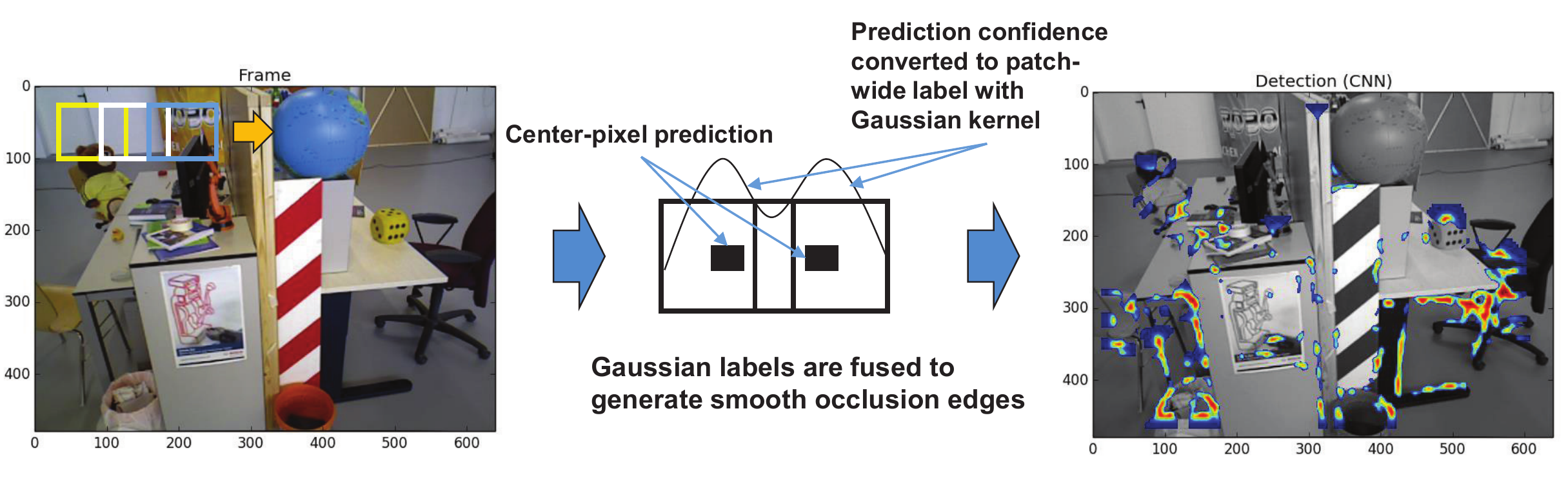}\vspace{-0pt}
  \caption{\textit{Post-processing at the testing phase involves collecting $32 \times 32$ overlapping patches with a constant stride from large frames; prediction confidence of a patch center pixel label is converted into a Gaussian kernel with Full Width at Half Maximum (FWHM); Gaussian labels are fused in a mixture model to generate smooth occlusion edges}}\vspace{0pt}
  \label{fig:testing}
\end{figure*}

Performance testing of CNN is done in both quantitative and qualitative manner with various input information as will be explained in Section~\ref{sec:exp}. For quantitative results, classification errors are computed based on the model's ability to predict label of the center pixels of a test patch collected from a frame captured in testing section of camera motion. The qualitative observations and visualization are made using a post-processing scheme as illustrated in Fig.~\ref{fig:testing}. In this scheme, classification confidence for a patch center pixels is collected from the softmax posterior distribution and it is extrapolated across the patch using a Gaussian distribution with Full Width at Half Maximum (FWHM). Such Gaussian kernels from overlapping patches are fused in a mixture model to generate smooth occlusion edges in the testing frame.

\section{Experiments and Quantitative Results}\label{sec:exp}
Different experiments are performed with different sets of input data for comparative evaluation. They are described below along with corresponding quantitative performance of the CNN model:
\begin{figure*}
    \centering
    \includegraphics[width=0.45\linewidth]{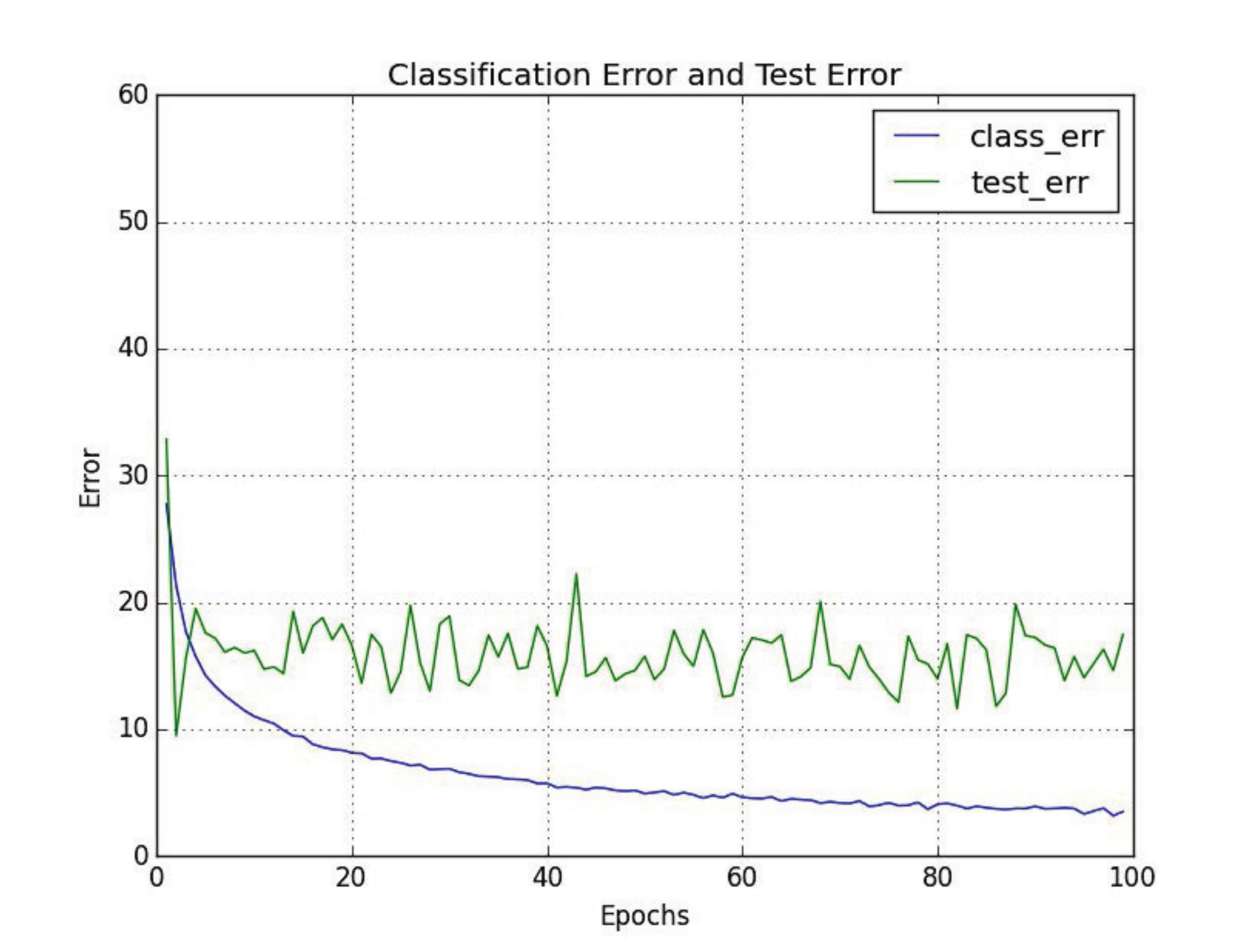}
    \includegraphics[width=0.45\linewidth]{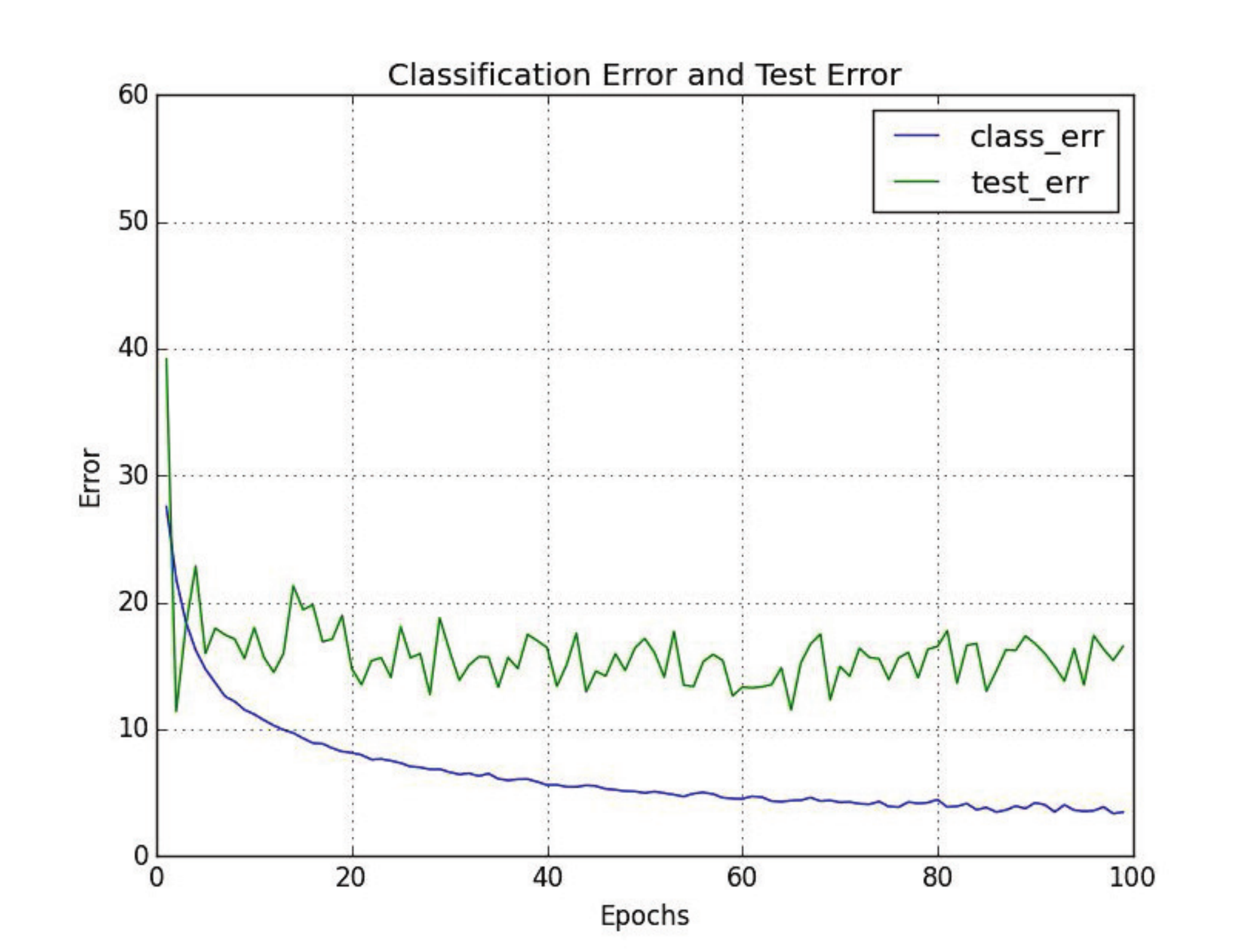}
    \caption{Training and testing error plots (for RGB-D input on left and RGB input on right) over various training epochs}
    \label{fig:error}
\end{figure*}

\textbf{RGB-D frame} The first set of experiments used single temporal frames of RGB-D data (i.e., $4$ channels). This task may seem rather straight forward as the depth information is directly available as one of the channels in the input data. However, majority of edges in the current frames are appearance edges and RGB channels clearly provide that information. Therefore, the task for the CNN model is to detect edges via automatic feature extraction and distinguishing occlusion edges from appearance edges. 

\textbf{RGB frame} The second set of experiments used single temporal frames of RGB data (i.e., $3$ channels). The goal here was to investigate if discriminative features exist and can be extracted by CNN from just RGB channels in order to classify patches into \textbf{Occlusion} and \textbf{No occlusion} edges. Ideally, without temporal information RGB channels may not carry a lot of occlusion information. However, occlusion information may remain in certain features such as shadows. Therefore, the objective is to investigate if such features can be recognized by a CNN to detect occlusion edges.  

\begin{table}
\begin{center}
\begin{tabular}{|c|c|p{1cm}|c|}
  \hline
  Channels & Overall error & False alarm & Missed detection \\
  \hline
  RGB-D & 15.7\% & 15.38\% & 43.59\% \\
  \hline
  RGB & 15.74\% & 15.42\% & 45.71\% \\
  \hline
\end{tabular}
\label{table:error_comparisons}
\end{center}
\caption{Occlusion detection performance of CNN with RGB-D and RGB inputs}
\end{table}

Numerical results are provided below for both of these cases. For training the CNN, $57,518$ training patches extracted from large image frames (collected in training section of the camera trajectory) are used. During testing, $1,271,002$ patches (collected in testing section of the camera trajectory) are used to provide quantitative performance data. Figure~\ref{fig:error} shows training and testing error plots (for both cases) over various epochs and specifically the training error graph clearly demonstrates that the training process does not saturate. This is due to the ReLU connections used in the CNN. As provided in Table 1, for both cases, false alarm performance is significantly better compared to missed detection performance. Numerically, overall error percentage is very close to false alarm rate as majority of the test example patches do not contain occlusion edges. Finally, as expected detection performance with RGB input is inferior (by ~$2\%$) to that of RGB-D input. However, false alarm rates are quite comparable. Overall, it is interesting to observe that performance degradation is not very large as input data changes as the depth (D) channel is removed.


\section{Qualitative Observations}\label{sec:qual}
\begin{figure*}[h]
  \centering\vspace{0pt}
  \includegraphics[width=1\textwidth]{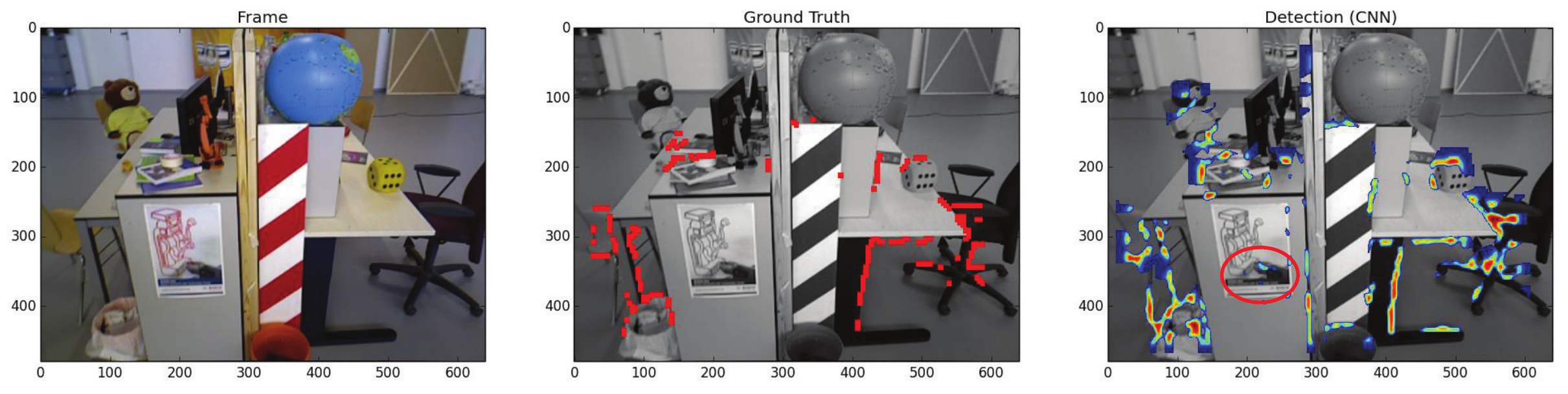}\vspace{-0pt}
  \caption{\textit{Occlusion detection performance on a test frame with RGB-D input and stride $4$; heat map shows the fused detection confidence (red-yellow-blue signifies high-medium-low; red circled region shows example of mistaking appearance edges as occlusion edges) }}\vspace{0pt}
  \label{fig:rgbd4}
\end{figure*}

\begin{figure*}[h]
  \centering\vspace{0pt}
  \includegraphics[width=1\textwidth]{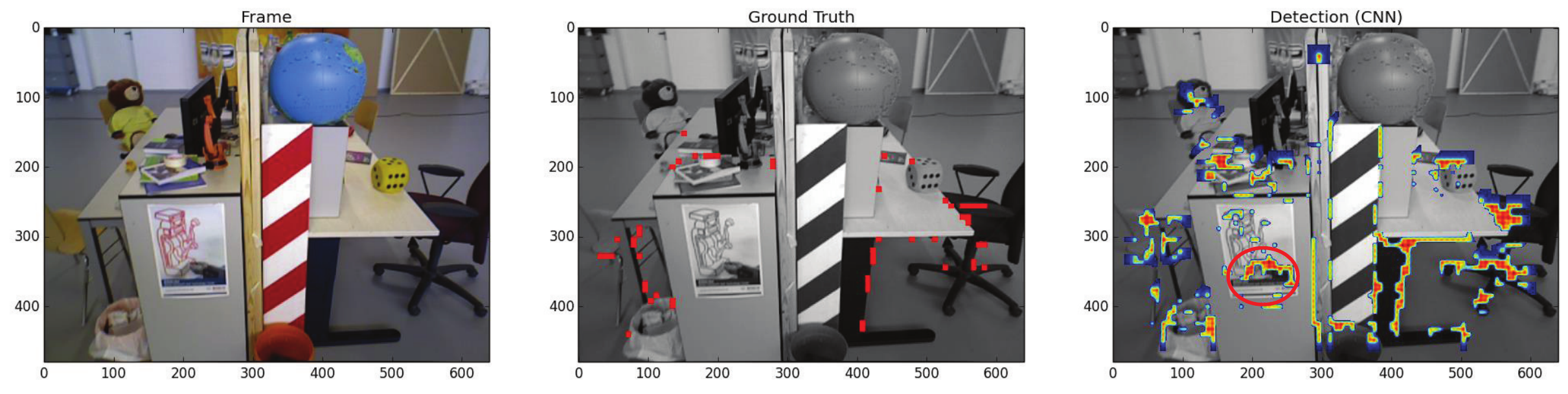}\vspace{-0pt}
  \caption{\textit{Occlusion detection performance on a test frame with RGB-D input and stride $8$; heat map description same as in Fig.~\ref{fig:rgbd4}; red circled region shows inferior performance compared to that of stride $4$}}\vspace{0pt}
  \label{fig:rgbd8}
\end{figure*}

This section presents qualitative results in order to understand the efficacy of the deep learning tools for occlusion edge detection and for robotics applications as a whole. Figures~\ref{fig:rgbd4} and~\ref{fig:rgbd8} show performances with RGB-D input with stride $4$ and $8$ (see Section~\ref{sec:testing} for details on strides) on a testing frame. As expected, occlusion edge generation is better with a lower value of stride as more information is available per pixel in this case. It can be noted in the marked regions (circled in red) in the figures that false detection of occlusion edges reduces with a lower value of stride. The trade-off lies in computational speed. With a lower value of stride, the frame processing time increases linearly with increase in number of test patches. Therefore, this trade-off has to be chosen properly for real-time robotics applications.

\begin{figure*}[h]
  \centering\vspace{0pt}
  \includegraphics[width=1\textwidth]{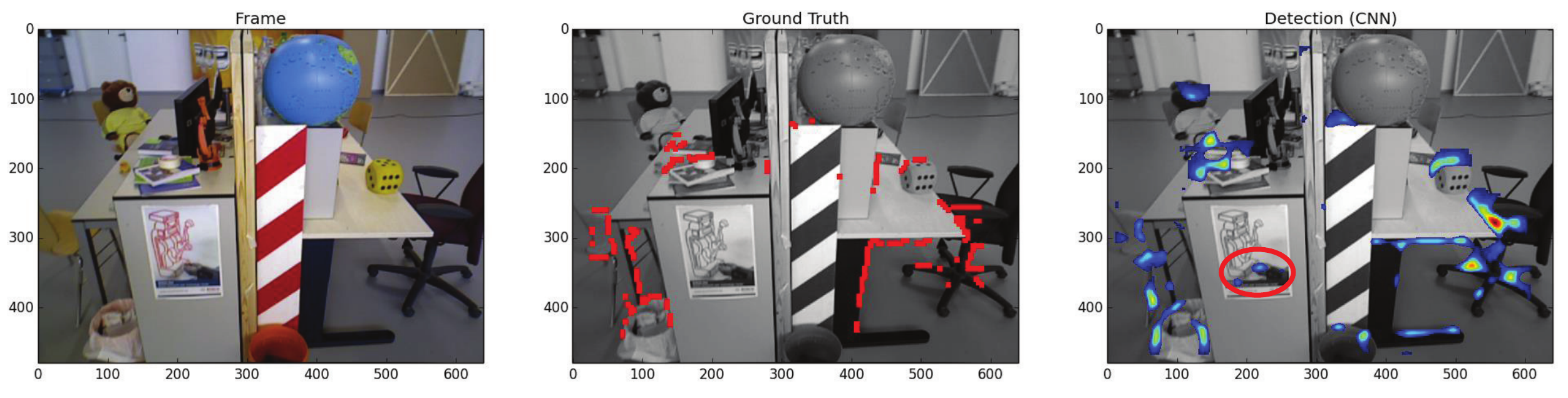}\vspace{-0pt}
  \caption{\textit{Occlusion detection performance on a test frame with RGB input and stride $4$; heat map description same as in Fig.~\ref{fig:rgbd4}; red circled region shows example of mistaking appearance edges as occlusion edges}}\vspace{0pt}
  \label{fig:rgb4}\vspace{-10pt}
\end{figure*}

\begin{figure*}[h]
  \centering\vspace{0pt}
  \includegraphics[width=1\textwidth]{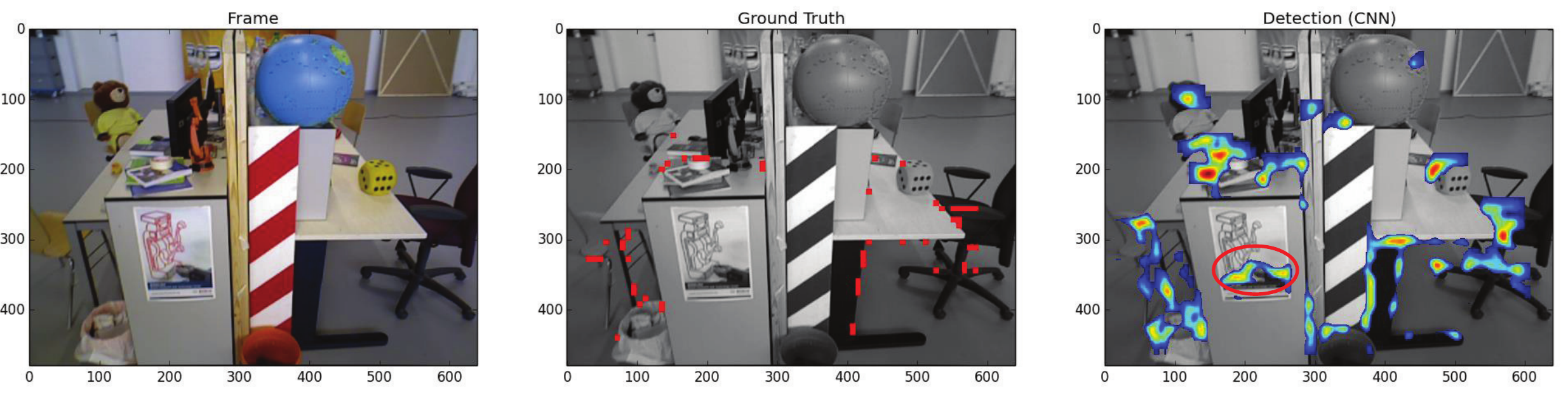}\vspace{-0pt}
  \caption{\textit{Occlusion detection performance on a test frame with RGB input and stride $8$; heat map description same as in Fig.~\ref{fig:rgbd4}; red circled region shows inferior performance compared to that of stride $4$}}\vspace{0pt}
  \label{fig:rgb8}\vspace{-10pt}
\end{figure*}

Figures~\ref{fig:rgb4} and~\ref{fig:rgb8} show performances with RGB input with stride $4$ and $8$ on the same testing frame and very similar observations can be made in this case as well. The heat maps also demonstrate decrease in detection confidence in this case compared to that of RGB-D input.  

\section{Conclusions and Future works}\label{sec:con}
In this study, we trained deep convolutional neural networks in a supervised manner in order to detect occlusion edges in RGB-D frames. The problem is formulated as a center-pixel classification problem for an image patch extracted from a larger frame. Apart from RGB-D inputs, experiments were performed to investigate the performance degradation associated with dropping the depth (D) channel. It is noted that although the missed detection rate increases slightly without depth data, the false alarm performance does not degrade significantly. A testing and post-processing scheme is developed to visualize the testing performance. The trade-off between high resolution patch analysis and frame-level computation time is discussed which is critical for real-time robotics applications. RGB-D and RGB frames lie on the two ends of the spectrum of input data information content. Therefore, investigations are currently being pursued with multiple time-frames of RGB input in order to extract structure from motion. Apart from this task, the other research directions are: (i) design of motion planning using decisions from CNNs and (ii) analysis of computation speed vs accuracy trade-off for real-time operation.



\bibliography{RGBD_DL}
\bibliographystyle{IEEEtran}

\end{document}